\documentclass{article}

\usepackage[preprint]{neurips_2020}

\usepackage[utf8]{inputenc} 
\usepackage[T1]{fontenc}    
\usepackage{hyperref}       
\usepackage{url}            
\usepackage{booktabs}       
\usepackage{amsfonts}       
\usepackage{nicefrac}       
\usepackage{microtype}      

\usepackage{enumitem}
\usepackage{amsmath}
\usepackage{graphicx}
\usepackage{subfigure}
\usepackage[square, comma, sort&compress, numbers]{natbib}

\usepackage{makecell}

\title{Self-Supervised Learning Aided Class-Incremental Lifelong Learning}

\author{%
  Song Zhang\\
  School of Electronics Engineering \\
  and Computer Science \\
  Peking University\\
  Beijing, CHN \\
  \texttt{songz@pku.edu.cn} \\
  \And
  Gehui Shen \\
  School of Electronics Engineering \\
  and Computer Science \\
  Peking University \\
  Beijing, CHN \\
  jueliangguke@pku.edu.cn \\
  \And
  Jinsong Huang \\
  School of Electronics Engineering \\
  and Computer Science \\
  Peking University \\
  Beijing, CHN \\
  huangjs@pku.edu.cn \\
  \AND
  Zhi-Hong Deng \\
  School of Electronics Engineering \\
  and Computer Science \\
  Peking University \\
  Beijing, CHN \\
  zhdeng@pku.edu.cn \\
}

\begin{document}

\maketitle

\begin{abstract}
Lifelong or continual learning remains to be a challenge for artificial neural network, as it is required to be both stable for preservation of old knowledge and plastic for acquisition of new knowledge. 
It is common to see previous experience get overwritten, which leads to the well-known issue of catastrophic forgetting, especially in the scenario of class-incremental learning (Class-IL). 
Recently, many lifelong learning methods have been proposed to avoid catastrophic forgetting.
However, models which learn without replay of the input data, would encounter another problem which has been ignored, and we refer to it as prior information loss (PIL).
In training procedure of Class-IL, as the model has no knowledge about following tasks, it would only extract features necessary for tasks learned so far, whose information is insufficient for joint classification.
In this paper, our empirical results on several image datasets show that PIL limits the performance of current state-of-the-art method for Class-IL, the orthogonal weights modification (OWM) algorithm.
Furthermore, we propose to combine self-supervised learning, which can provide effective representations without requiring labels, with Class-IL to partly get around this problem.
Experiments show superiority of proposed method to OWM, as well as other strong baselines.
\end{abstract}

\section{Introduction}
In recent years, deep neural networks have shown remarkable performance in a wide variety of individual tasks, and even surpass human experts in certain fields. 
However, humans and animals are better at continually acquiring, fine-tuning and transferring knowledge and skills throughout their lifetime, which benefits from a good balance between synaptic plasticity and stability \citep{Abraham2005Memory}.
In a lifelong learning scenario, an intelligent system requires sufficient plasticity to integrate novel information and stability to prevent significantly interfering with consolidated knowledge.
For machine learning and neural network models, lifelong learning represents a long-standing challenge.
Since continual acquisition of information from non-stationary data distributions generally leads to catastrophic
forgetting \citep{Mccloskey1989Catastrophic,Robins1993Catastrophic,French1993Catastrophic}, in which new knowledge overwrites old knowledge, leading to a quick, pronounced drop of the performance on previous tasks. 
Catastrophic forgetting has been a key obstacle for deep neural networks to learn sequentially. 

In the setting of lifelong learning, only data from the current task is available during training and the tasks are assumed to be clearly separated.
In general, there are three main types of lifelong learning scenarios \citep{van2019Three,Hsu2018Re}, based on whether task identity is provided at test time and, if not, whether task identity must be inferred.
They are task-incremental learning (Task-IL), domain-incremental learning (Domain-IL) and class-incremental learning (Class-IL).
In this paper, we focus on Class-IL since it is the most challenging one.
Class-IL includes the common problem of incrementally learning new classes, in which the classes of each task are disjoint and the model must be able to both solve each task seen so far and infer which task they are presented with, in other words, distinguish all the classes.

Recently, there have been multiple attempts to mitigate catastrophic forgetting \citep{coop2013ensemble,Goodrich2014Unsupervised,gepperth2016bio,Fernando2017PathNet,Lee2017Lifelong,Lopez2017Gradient}.
And many state-of-the-art methods \citep{Li2016Learning,Kirkpatrick2017Overcoming,Zenke2017Continual} in other two scenarios are not capable to handle Class-IL \citep{van2019Three,Hsu2018Re}.
Basically, there are there types of strategies which work in Class-IL scenario: \emph{exemplar replay} \citep{Rebuffi2016iCaRL,Nguyen2017Variational}, \emph{generative replay} \citep{shin2017continual,Kamra2017Deep} and \emph{gradient projection} \citep{he2018overcoming,zeng2019continual}.
\emph{Exemplar replay} is a simple strategy to suppress catastrophic
forgetting by selecting and storing a subset of input data under certain constraint. 
However, such methods violate the purpose of lifelong learning since data of previous tasks should be unavailable. 
An alternative is to approximate the data distribution of previous tasks with a separate generative model for replay.
The performance of \emph{generative replay} relies on the complexity of training data, as the generator is also trained in a incremental manner, which is much difficult than that in joint manner \citep{Timothee2019Generative}.
The last kind of method retains previously learned knowledge
by keeping the old input-output mappings stable. 
Orthogonal Weights Modification (OWM) \citep{zeng2019continual} is a typical method and can be seen as the state-of-the-art method for Class-IL scenario.

\begin{figure}[htbp]
\label{animal}
\centering
\includegraphics[width=10cm,height=5cm]{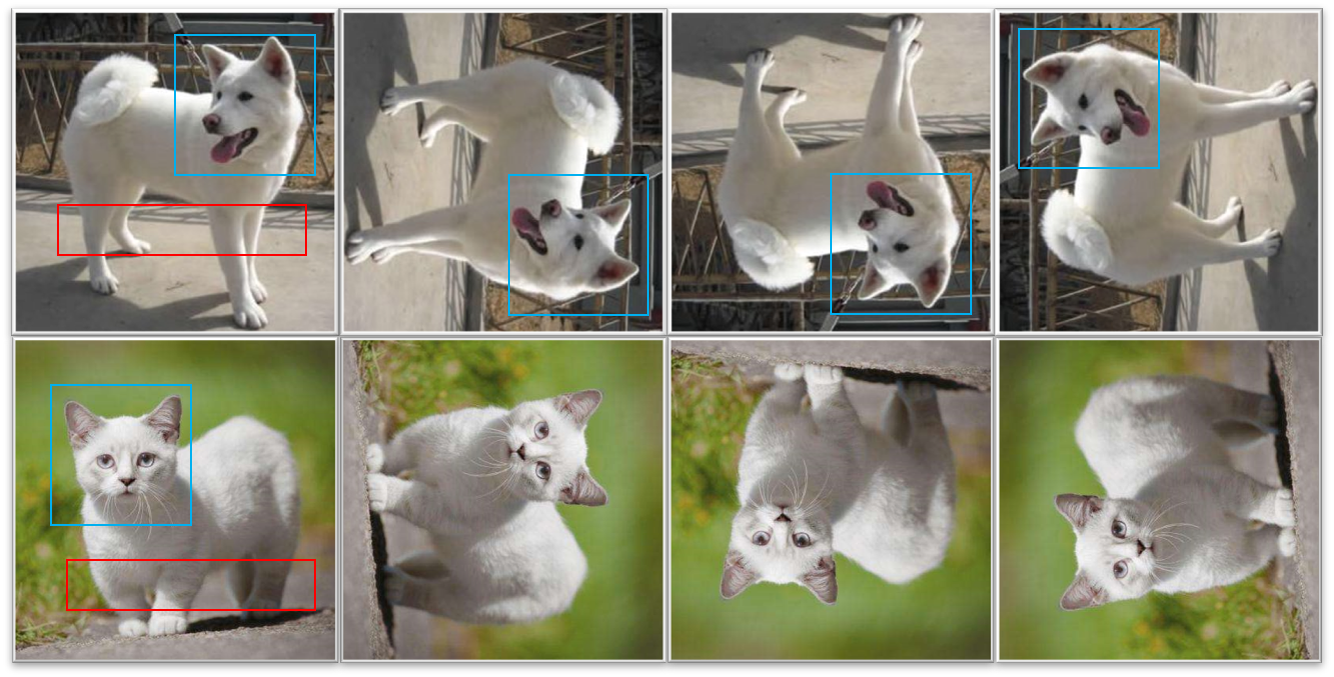}
\caption{Relying on features learned in previous tasks, e.g., number of legs (red box), model might confuse previous classes(e.g., dog) with current ones(e.g., cat). Proxy task of prediction rotation requires modeling distinctive shapes, e.g., head (blue box) on each task, which helps classification. }
\end{figure}

Though OWM showed remarkable performance on different datasets, it is confronted with another problem in addition to catastrophic forgetting, compared with the other two strategies.
Given images of one category, humans could observe and extract multiple meaningful features based on their common sense, which the lifelong learning models are not equipped with.
Their optimization is motivated by the objective function of classification, which would not demand for features unnecessary for current task, even though they might be required in following tasks.
We refer to this phenomenon as prior information loss (PIL), as the model has no idea of the features necessary for joint classification as prior knowledge.
For example, model can easily classify dog and bird by learning to count legs, and when cat appears in next task, more attributes would be required (Figure \ref{animal}).
However without replay of previous inputs, it is unable to mine extra information of previous classes as they are not available anymore.
As it goes on, more features would be selected, which are only extracted for current and following classes and the missing parts affect the joint classification accuracy.

To solve the problem adequately, a straight way is combining OWM with \emph{generative replay}.
However, due to its own difficulty, \emph{generative replay} has negative effect on OWM \citep{shen2020generative}.
Without opportunity to train on current data later, we need to extract features in an unsupervised way as more as possible for backup.
In this paper, we propose to exploit self-supervised learning (SSL) as an alternative substitution, which can
provide effective representations without labels, to partly solve the problem.
Specifically, we train the model to classify and predict self-supervised labels in the manner of multi-task learning.
As the same representations are shared for predicting original labels and self-supervised signals, the knowledge acquired is enriched by that learned via SSL.
In following tasks, we have more informative features of previous classes for distinguishment, while the improvement of accuracy hinges on the relevance between the originally missing features and selected self-supervised signals.

Our contributions are two-fold: 
Firstly, we first propose the problem of prior information loss (PIL), another obstacle besides the catastrophic forgetting, which is applicable to models without input replay, among which we regard OWM as backbone. 
We also design experiments on different datasets to show its effect, which restricts the performance of OWM with a significant margin. 
Secondly, we combine self-supervised learning with Class-IL to make up for this margin, which is simple but effective manner to extract more valuable features in each task. 
Experimental results on several datasets show that proposed method gets steady improvement over OWM, and performs better than state-of-the-art methods of other strategies.

\section{Related Work}
Catastrophic forgetting firstly explored in 1980s and 1990s \citep{Mccloskey1989Catastrophic,Robins1993Catastrophic,French1993Catastrophic}, has attracted more and more attentions in recent years and lifelong learning has also been extended to unsupervised settings \citep{rao2019continual} and semi-supervised settings \citep{smith2019unsupervised}.
For supervised lifelong learning, many approaches have been presented. 
While not exhaustive, we roughly divide them into five strategies, including \emph{task-specific}, \emph{regularization}, \emph{exemplar replay}, \emph{generative replay} and \emph{gradient projection}.

The motivation of \emph{task-specific} methods is that previously acquired knowledge can be preserved by only optimizing part of the network. 
Several papers use this strategy, with different approaches for selecting the parts of the network for each task, e.g., dynamically expanding \citep{Rusu2016Progressive}, randomly selecting \citep{Masse2018Alleviating}, evolutionary algorithms \citep{Mallya2018PackNet} or attention \citep{Serr2018Overcoming}.
However, these methods are only be applied to Task-IL, in which task identity is required to select corresponding parts during test.

\emph{Regularization} methods, as well as the following methods, do not need task identity as the entire network is shared among different tasks.
\emph{Regularization} methods avoid interfering with prior knowledge by adding constraints to the weights updates.
By estimating the importance of each weight for previous tasks, different penalty of changes are imposed to them, like Elastic Weight Consolidation (EWC) \citep{Kirkpatrick2017Overcoming}, Synaptic Intelligence (SI) \citep{Zenke2017Continual} and Memory Aware Synapses (MAS) \citep{aljundi2018memory}.
While these methods are effective on Task-IL and Domain-IL scenarios, they show poor performance on Class-IL scenario \citep{Hsu2018Re,van2019Three}.

Though violating the purpose of lifelong learning, \emph{exemplar replay} methods provide a strong baseline, by storing data from previous tasks, called "exemplars".
A typically example is iCaRL[\citep{Rebuffi2016iCaRL}], which uses neural network for feature extraction and nearest mean of exemplars for classification.
It manages an exemplar set with a fixed size and combines it with current inputs.

\emph{Generative replay} methods gain more support from biological evidence, which suggests that the hippocampus is more than a simple experience replay buffer and reactivation of the memory traces yields rather flexible outcomes \citep{ramirez2013creating}.
The strategy is to train a separate generative model sequentially, which approximates the data distributions of all previous tasks.
When presented with a new task, generated samples are interleaved with new data to update the generator and classifier.
The generative adversarial networks (GANs) framework\citep{goodfellow2014generative} are often utilized, like generative replay (DGR) (\citep{shin2017continual}) and Dynamic Generative Memory (DGM) (\citep{ostapenko2019learning}). 
And the complementary learning systems (CLS) theory \citep{Randall2002Hippocampal,Kumaran2016What} can also be modelled \citep{Kamra2017Deep}.

Finally, \emph{gradient projection} methods try to retain learned knowledge by keeping the mappings trained on different tasks fixed, and learn new ones while avoiding conflicts with them.
\citep{zeng2019continual} achieves this goal by orthogonal weights modification (OWM), in which weights are only allowed to be modified in the direction orthogonal to the subspace spanned by all previous inputs.
OWM shows good ability for overcoming catastrophic forgetting, and exhibits superiority in comparison with other methods in Class-IL scenario.

\subsection{Self-Supervised Learning}
Self-supervised learning (SSL) is applied to improving representations when labeled data is expensive and impractical to scale up.
Recent literatures \citep{ji2018invariant,oord2018representation,hjelm2018learning,henaff2019data,zhang2019aet}, have proven that semantic representations can be learned by predicting labels obtained from the input signals without any human annotations.
Several proxy tasks with different signals have been proposed, including relative position of image patches \citep{doersch2015unsupervised}, colorization after shuffle \citep{larsson2017colorization}, image rotations \citep{gidaris2018unsupervised} and so forth \citep{dosovitskiy2015discriminative,noroozi2016unsupervised}.

While originally focused on unsupervised learning, it has also been extended to semi-supervised learning and many methods have been state-of-the-art \citep{dosovitskiy2015discriminative,zhai2019s4l}.
Recently, there are also many attempts for related tasks, e.g., adversarial generative networks \citep{chen2019self}.
However, while combined with fully-supervised learning, SSL is generally used for data augmentation, and has negative effect on accuracy, though benefits on generality, robustness and uncertainty \citep{hendrycks2019using}.

Commonly, original task and proxy task are combined in a multi-task learning strategy, while sharing feature representations.
\citep{Lee2019Rethinking} proposed not to assign the same label to all augmented samples of the same source.
Instead, learning the joint distribution of the original labels and self-supervised signals of augmented samples, and combining the predictions from different augmented samples help improve the performance of supervised learning.

\section{Methodology}

\subsection{Formulation and Analysis}
\label{analysis}

In this paper, we study the problem of Class-IL, which is characterized by $ T $ tasks of supervised learning, with dataset $D={\{D_t\}}_{t=1}^T$.
Dataset of each task $ D_t $ consists of $N_t$ labeled training samples of corresponding classes, i.e., $D_t=\{x_{t,j}, y_{t,j}\}_{j=1}^{N_T}$, where $y_{t,j} \in C_t$ and $C_t$ is the class set of task $t$.
Note that class sets of different tasks are disjoint and the model is trained sequentially which means when presented with task $t$, only current training set is available.
After training on all tasks, the model is evaluated on test data of all classes that have been learned. 

Neural networks for Class-IL scenario all share the same output layer for different tasks, as task identify is not provided during test.
For the convenience of discussion, we divide these models into two parts: the classifier $C$ which corresponds to the last fully-connected layer, and the feature extractor $E$ which corresponds to all previous layers.
After training on task $t$, the output of such models is:
\begin{equation}\nonumber
p(y_{\leq t}|x_{\leq t},\theta_t,\phi_t)=C_t(E_t(x_{\leq t},\theta_t),\phi_t)
\end{equation}
where $\theta$ and $\phi$ mean the parameters of $E$ and $C$ respectively.
From this perspective, previous works for alleviating catastrophic forgetting can be divided into two categories, despite of the difference of implementation method:
\begin{itemize}[leftmargin=*]
\item When presented with a new task $t$, replay-based methods both modify their representations of previous inputs to distinguish between them and current classes, i.e. $E(x_{<t},\theta_t) \neq E(x_{<t},\theta_{t-1})$.
Therefore, the classifier would also have to relearn the mapping from feature space to classes, based on its latest distribution , sampled from both current real data and exemplars or generated pseudo-data.
\item On the other hand, \emph{regularization} and \emph{gradient projection} methods try to keep the features extracted from previous data stable, i.e. $E(x_{<t},\theta_t)\approx E(x_{<t},\theta_{t-1})$.
However, previous representations $E_{<t}(x_{<t})$ are learned to distinguish classes that have been seen at that time.
Without update, their information is insufficient for the following classification requirements, which we refer to as prior information loss (PIL).
\end{itemize}
For formalization of PIL, we define $A(x)$ to be the universal set of disentangled features that can be extracted from sample $x$.
In the scenario of joint classification, there might be $K$ different subsets of $A(x)$ that satisfy this task without redundancies, $\{N_k(x)\}_{k=1}^K$, and we refer to them as \emph{eligible subsets}.
Ignoring the capacity limit of $E$, we obtain $ E(x,\theta) $ covering at least one eligible subsets, e.g., $ E(x,\theta) \supseteq N_1(x) $.
When it turns to multiple sequential tasks, ideally, we also obtain $ E(x_t,\theta_t)\supseteq N_1(x_t)$ while training on task $t$ and retain it till the end.
However, except for the effect of catastrophic forgetting, another problem is that the model has no idea about the eligible subsets, while it might be common sense for human beings.
When training on the first task, we obtain $ E(x_1,\theta_1) $ and there exists eligible subsets covering it, e.g., $N_1(x_1) \supseteq E(x_1,\theta_1)$.
Generally, they are not equal unless it is as difficult as joint classification.
Suppose catastrophic forgetting does not occur in following procedure. When training on task t, more attributes are added into $ E(x_t,\theta_t)$ while that of old classes remains unknown, i.e. for $i<t$, $E(x_i,\theta_t)=E(x_i,\theta_i)$.
Without loss of generality, we assume these added attributes are also in $ N_1(x_t) $ except the redundant ones.
Finally, we obtain $E(x_t,\theta_T)=E(x_t,\theta_t)$ for each task $t$ and there exists difference set $D(x_t)=N_1(x_t)-E(x_t,\theta_T)$, which affects the classification accuracy.

\subsection{Combination with self-supervised learning}
PIL seems intractable when the size of $A(x)$ is large and we can only obtain a limited number of them. 
To a certain extent, we can make up for this problem, by imposing prior information about which attributes might be important.
We propose that self-supervised learning (SSL) \citep{larsson2017colorization,gidaris2018unsupervised} is a simple but effective way to achieve this goal.
Since in the training procedure of Class-IL scenario, representation learning without any information of following tasks, can also seen as a form of unsupervised learning, which is the strength of SSL.
For a selected proxy task of SSL with $\{f_m\}_{m=1}^M$ as the set of transformations and given a transformed sample $f_m(x_t)$, the model learns to predict which transformation is applied.
Supervised classification learning and self-supervised learning are trained at the same time during each task, in a multi-task manner.
Besides, they share the same feature extractor $E$ while SSL has a separate output layer $F$.
Let $L_{ce}$ be the cross-entropy function, the overall loss function is as follow:
\begin{equation}\nonumber
L_t(x_t,y_t;\theta_t,\phi_t,\sigma_t)=L_{ce}(C(E(x_t;\theta_t);\phi_t),y_t)+\frac{\alpha_t}{M} \sum_{m=1}^M L_{ce}(F(E(f_m(x_t);\theta_t);\sigma_t),m)
\end{equation}
where $\sigma$ is the parameter of $F$ and $\alpha_t$ is a hyper-parameter which decreases when stitching to next task.
As $E$ is shared between the two parts, we now have $E(x_t,\theta_t)=E_{sl}(x_t,\theta_t) \cup E_{ssl}(x_t,\theta_t)$, where $E_{sl}(x_t,\theta_t)$ and $E_{ssl}(x_t,\theta_t)$ represent features required by supervised learning and self-supervised learning respectively.
Improvement on final classification accuracy relies on the intersection, $D(x_t)\cap E_{ssl}(x_t,\theta_t)$.
Note that data augmentation is not applied as there is no need to improve upon the generalization ability in the case of under-fitting.
Besides, we need not to alleviate catastrophic forgetting of SSL, as its output layer $F$ is not used for prediction during test.

\subsection{Training with OWM}
We optimize $F$ with gradient computed by back propagation (BP) while optimize $E$ and $C$ with OWM algorithm \citep{zeng2019continual}.
Here we give a brief introduction of OWM algorithm.
For a FC layer with weight matrix $W$, the key to overcome catastrophic forgetting in sequential learning, is the orthogonal projector $P$ defined on its input space for learned tasks.
In general, the projector is defined as $P_t=I-A_t(A_t^TA_t+ \alpha I)^{-1}A_t^T$ for training task $t$.
Matrix $A_t$ consists of all trained input vectors spanning
the input space that have been trained as its columns, e.g.,$A_t=[a_1,...a_{t-1}]$, and $\alpha$ donates a small constant for avoiding the ill-conditioning problem in the matrix-inverse operation.
Moreover, $P_t$ can be updated recursively based on current input and $P_{t-1}$, which greatly reduces the complexity of calculation.

Given $P_t$ for current task $t$ and gradient computed by back propagation $\Delta W_t$, the gradient is modified with $\Delta W_t^{OWM}=\Delta W_t P_t$.
For input of previous tasks $a_{<t}$, which is spanned by $A_t$, we have:
\begin{align*}
\nonumber  W_t a_{<t} &= ( W_{t-1} + \eta \Delta W_t^{OWM} ) a_{<t} \\
\nonumber &=W_{t-1}a_{<t}+\eta \Delta W_t P_t a_{<t} \\
\nonumber &=W_{t-1}a_{<t}+\eta \Delta W_t[I-A_t(A_t^TA_t+ \alpha \nonumber I)^{-1}A_t^T]a_{<t} \\
\nonumber &\approx W_{t-1}a_{<t} \\
\end{align*}
which means that the output is approximately invariable after training on new task.
For mini-batch training, $P_t$ can also be updated successively after current task $t$ is completed.
Each time we calculate the mean of the inputs for the $i^{th}$ batch $\overline{x}_{t,i}$, then $P_t$ can be calculated iteratively as follow:
\begin{equation}\nonumber
k_{t,i}=P_{t,i-1} \overline{x}_{t,i} / [1+{\overline{x}_{t,i}}^T P_{t,i-1} \overline{x}_{t,i}]
\end{equation}
$$P_{t,i}=P_{t,i-1}-k_{t,i} {\overline{x}_{t,i}}^T P_{t,i-1}$$
in which $P_{t,0}=P_{t-1}$ and after iteration, we have $P_t=P_{t,N_t}$.
This algorithm can also be extended to convolution neural networks (CNN).

\section{Experiments}
In this section, we conduct several experiments on Class-IL scenario, to show the impact of PIL, and applicability of our proposed method, which will be referred to as OWM+SSL in the following text.
In section \ref{sec-settings}, we introduce our datasets, neural network structure and setting of baselines.
In section \ref{sec-e1}, we evaluate our method comparing with several state-of-the-art baselines of different strategies.
In section \ref{sec-e2}, we explore the upper bound of OWM by training with distillation of joint pre-trained representations, to observe its performance without PIL.
Finally in section \ref{sec-e3}, we explore how different SSL usages and choices of proxy tasks affect the result.

\subsection{Settings}\label{sec-settings}
We use three image datasets collected from real world:
\begin{itemize}[leftmargin=*]
\item SVHN: digit images from house numbers, contains 32x32 colour images in 10 classes. There are 73257 training images and 26032 test images.
\item CIFAR10: consists of 60000 32x32 colour images in 10 classes of common objects, with 6000 images per class. There are 50000 training images and 10000 test images.
\item CIFAR100: just like the CIFAR-10, except it has 100 classes containing 600 images each. There are 500 training images and 100 testing images per class.
\end{itemize}
And we conduct Class-IL experiments with 5 tasks for SVHN and CIFAR10, 2/5/10 tasks for CIFAR100.
For each dataset, a subset of test set in randomly selected for validation, while the rest used for final test set.
The neural network is composed of a 3-layer CNN with 64, 128, 256 2$\times$2 filters and 3-layer MLP with 1000 hidden units.
The number of filters are doubled for CIFAR100.
We set learning rate as 0.1 for SVHN and CIFAR10, 0.05 for CIFAR100.
The value of $\alpha_t$ is calculated as follow:
\begin{equation}\nonumber
\alpha_t=\frac{T-t}{T-1}alpha
\end{equation}
where alpha is 5.0 for SVHN and CIFAR10, 0.75 for CIFAR100.

We compared the following methods, which shares the same architecture of neural network:
\begin{itemize}[leftmargin=*]
\item \textbf{EWC and MAS}: classical regularization methods.
\item \textbf{iCaRL}: a strong baseline of exemplar replay, with a memory budget of 2000.
\item \textbf{DGR}: a separate AC-GAN \citep{Wu2018Memory} is trained for generative replay.
The generator and discriminator have 3 deconvolution and convolution layer respectively.
The replayed images were labeled with the most likely class predicted by a copy of the main model stored after training on the previous task.
\item \textbf{DGM}: utilize neural masking to realize a conditional GAN with learnable connection plasticity.
\item \textbf{OWM}: state-of-the-art method for Class-IL scenario, as well as the basis of our method.
\end{itemize}

\subsection{Classification Results}
\label{sec-e1}
In this subsection, we display the test accuracies of all the 
datasets after all tasks are learned in Table \ref{table-e1} and some results are quoted from \citep{shen2020generative}.
For proposed OWM combined with SSL (OWM+SSL), we exploit predicting the rotation of images \citep{gidaris2018unsupervised} as proxy task.
As conclusion in \citep{Hsu2018Re,van2019Three}, EWC totally fails in Class-IL scenario.
Its correct predictions are almost all from the last task, which means that previously learned knowledge are totally forgetten.
In most setting, iCaRL and DGR show comparable results, except in 5-tasks CIFAR10.
In tasks of CIFAR100, performance of iCaRL and DGR both fall faster as the number of task increases, compared with OWM.
Since it is getting more difficult to approximate data distribution of all learned classes for them.
Compared with other baselines, original OWM has shown a clear superiority in all settings.

\begin{table}[htbp]
  \caption{Results after all tasks are learned.}
  \label{table-e1}
  \centering
  \resizebox{\textwidth}{15mm}{
  \begin{tabular}{llllll}
    \toprule
    \makecell[c]{Method} & \makecell[c]{SVHN} & \makecell[c]{CIFAR10} & \makecell[c]{CIFAR100} & \makecell[c]{CIFAR100} & \makecell[c]{CIFAR100} \\
     & \makecell[c]{(5 tasks)} & \makecell[c]{(5 tasks)} & \makecell[c]{(2 tasks)} & \makecell[c]{(5 tasks)} & \makecell[c]{(10 tasks)} \\
    \midrule
    \makecell[c]{EWC \citep{Kirkpatrick2017Overcoming}} & \makecell[c]{$12.25 \pm 0.13$} & \makecell[c]{$18.53 \pm 0.11$} & \makecell[c]{$24.31 \pm 0.63$} & \makecell[c]{$12.53 \pm 0.69$} & \makecell[c]{$7.56 \pm 0.25$} \\
    \makecell[c]{MAS \citep{aljundi2018memory}} & \makecell[c]{$18.11 \pm 1.80$} & \makecell[c]{$20.25 \pm 1.54$} & \makecell[c]{$29.04 \pm 0.92$} & \makecell[c]{$14.35 \pm 0.22$} & \makecell[c]{$8.44 \pm 0.27$} \\
    \makecell[c]{iCaRL \citep{Rebuffi2016iCaRL}} & \makecell[c]{$67.91 \pm 0.84$} & \makecell[c]{$57.66 \pm 0.86$} & \makecell[c]{$36.09 \pm 1.34$} & \makecell[c]{$30.07 \pm 0.98$} & \makecell[c]{$15.93 \pm 1.02$} \\
    \makecell[c]{DGR \citep{shin2017continual}} & \makecell[c]{$67.50 \pm 0.83$} & \makecell[c]{$22.39 \pm 0.83$} & \makecell[c]{$36.48 \pm 0.61$} & \makecell[c]{$25.52 \pm 0.46$} & \makecell[c]{$15.23 \pm 1.42$} \\
    \makecell[c]{DGM \citep{ostapenko2019learning}} & \makecell[c]{$72.68 \pm 1.49$} & \makecell[c]{$50.53 \pm 0.57$} & \makecell[c]{$28.23 \pm 0.72$} & \makecell[c]{$25.43 \pm 0.24$} & \makecell[c]{$17.43 \pm 1.25$} \\
    \makecell[c]{OWM \citep{zeng2019continual}} & \makecell[c]{$74.32 \pm 0.66$} & \makecell[c]{$54.89 \pm 0.73$} & \makecell[c]{$41.46 \pm 0.44$} & \makecell[c]{$34.39 \pm 0.41$} & \makecell[c]{$30.79 \pm 0.56$} \\
    \makecell[c]{\textbf{OWM+SSL (ours)}} & \makecell[c]{$\mathbf{82.08} \pm 0.72$} & \makecell[c]{$\mathbf{59.20} \pm 1.44$} & \makecell[c]{$\mathbf{42.23} \pm 0.22$} & \makecell[c]{$\mathbf{35.12} \pm 0.21$} & \makecell[c]{$\mathbf{31.76} \pm 0.68$} \\
    \bottomrule
  \end{tabular}}
\end{table}

Based on OWM, proposed OWM+SSL brings steady improvement and it is especially obvious on tasks of SVHN and CIFAR10.
In these settings, as there are relatively less classes in each task, the phenomenon of PIL is more serious, which can lead to greater decline in test accuracy.
At the same time, the amount of classes for joint testing is relatively small, which demands less feature for distinguishment.
Thus, the proportion of the supplement from SSL is larger, which contributes to classification.

When it turns to fine-grained classification on CIFAR100, much more detailed attributes are needed, most of which are not necessary for predicting the rotations of images.
While single proxy task help extracting more useful features, it is not enough to hold a considerable proportion of the missing parts.
In other words, there is still much room for improvement.
To tackle with this problem, perhaps the combination of different proxy tasks is a feasible option.

\subsection{Upper Bound of OWM}
\label{sec-e2}
In this subsection, we explore the upper bound of OWM without PIL.
First, we pre-train a model with the same structure in joint learning scenario, and we stored its feature extractor $E^*$ as the teacher model.
Then we train a model in Class-IL scenario using OWM, and during each task, the model learns to extract all the necessary features from $E^*$ by distillation, while learning to classify.
The loss function is as follow:
\begin{equation}\nonumber
L_t(x_t,y_t;\theta_t,\phi_t)=L_{ce}(C(E(x_t;\theta);\phi_t),y_t)+\lambda L_{mse}(E(x_t,\theta_t),E^*(x_t,\theta^*))
\end{equation}
where $L_{mse}$ represents the mean square error and $\lambda$ is a hyperparameter.
In such setting, the model is able to capture the same features with $E^*$, thus no PIL.
Compared with joint learning, the decline of classification is attributed to catastrophic forgetting and information loss of distillation.
Here we approximate it as the the upper bound of OWM, though the bound should be higher if there is no loss in distillation or the model can learn these representations by itself.

\begin{table}[htbp]
  \caption{Results of OWM with feature distillation.}
  \label{table-e2}
  \centering
  \begin{tabular}{l|l|l|lll}
    \toprule
    \makecell[c]{Method} & \makecell[c]{SVHN} & \makecell[c]{CIFAR10} & \makecell[c]{CIFAR100} & \makecell[c]{CIFAR100} & \makecell[c]{CIFAR100} \\
     & \makecell[c]{(5 tasks)} & \makecell[c]{(5 tasks)} & \makecell[c]{(2 tasks)} & \makecell[c]{(5 tasks)} & \makecell[c]{(10 tasks)} \\
    \midrule
    \makecell[c]{OWM \citep{zeng2019continual}} & \makecell[c]{$74.32$} & \makecell[c]{$54.89$} & \makecell[c]{$41.46$} & \makecell[c]{$34.39$} & \makecell[c]{$30.79$} \\
    \makecell[c]{OWM+SSL} & \makecell[c]{$79.54$} & \makecell[c]{$57.86$} & \makecell[c]{$41.97$} & \makecell[c]{$35.05$} & \makecell[c]{$31.46$} \\
    \makecell[c]{OWM+FD} & \makecell[c]{$87.26$} & \makecell[c]{$70.59$} & \makecell[c]{$49.99$} & \makecell[c]{$47.21$} & \makecell[c]{$45.19$} \\
    \midrule
    \makecell[c]{Joint Training} & \makecell[c]{$91.50$} & \makecell[c]{$77.51$} & \multicolumn{3}{c}{$52.29$} \\
    \bottomrule
  \end{tabular}
\end{table}

In this experiment, We set $\lambda=300$ for SVHN and CIFAR10, and $\lambda=100$ for CIFAR100 to rescale the losses.
The results are shown in Table \ref{table-e2}.
We report the classifications accuracies after joint training, which can be seen as the upper bound for lifelong learning with the same neural network.
The models are then used as the teacher model for OWM with feature distillation (OWM+FD).
There is an evident gap between OWM and OWM+FD in all settings, which shows the significant impact of PIL.
While the gap between OWM+FD and the teacher model is less obvious.
Thess results proves that OWM has achieved relative success in alleviating catastrophic forgetting, however is restricted by PIL to a greater extent, and SSL helps make up part of it.

\begin{figure}[htbp]
\centering
\subfigure[SVHN (5 tasks)]{
\includegraphics[width=4cm]{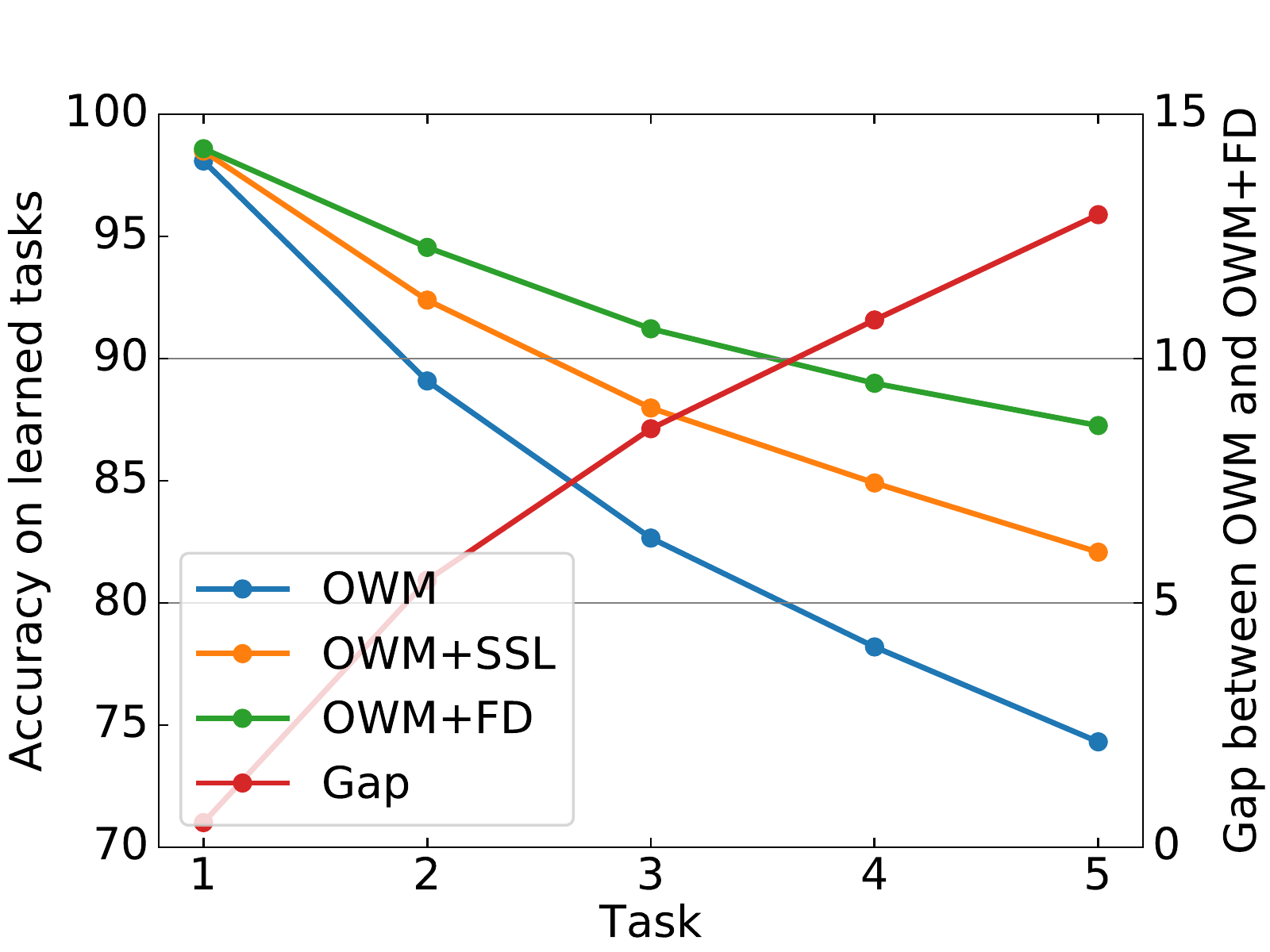}
}
\quad
\subfigure[CIFAR10 (5 tasks)]{
\includegraphics[width=4cm]{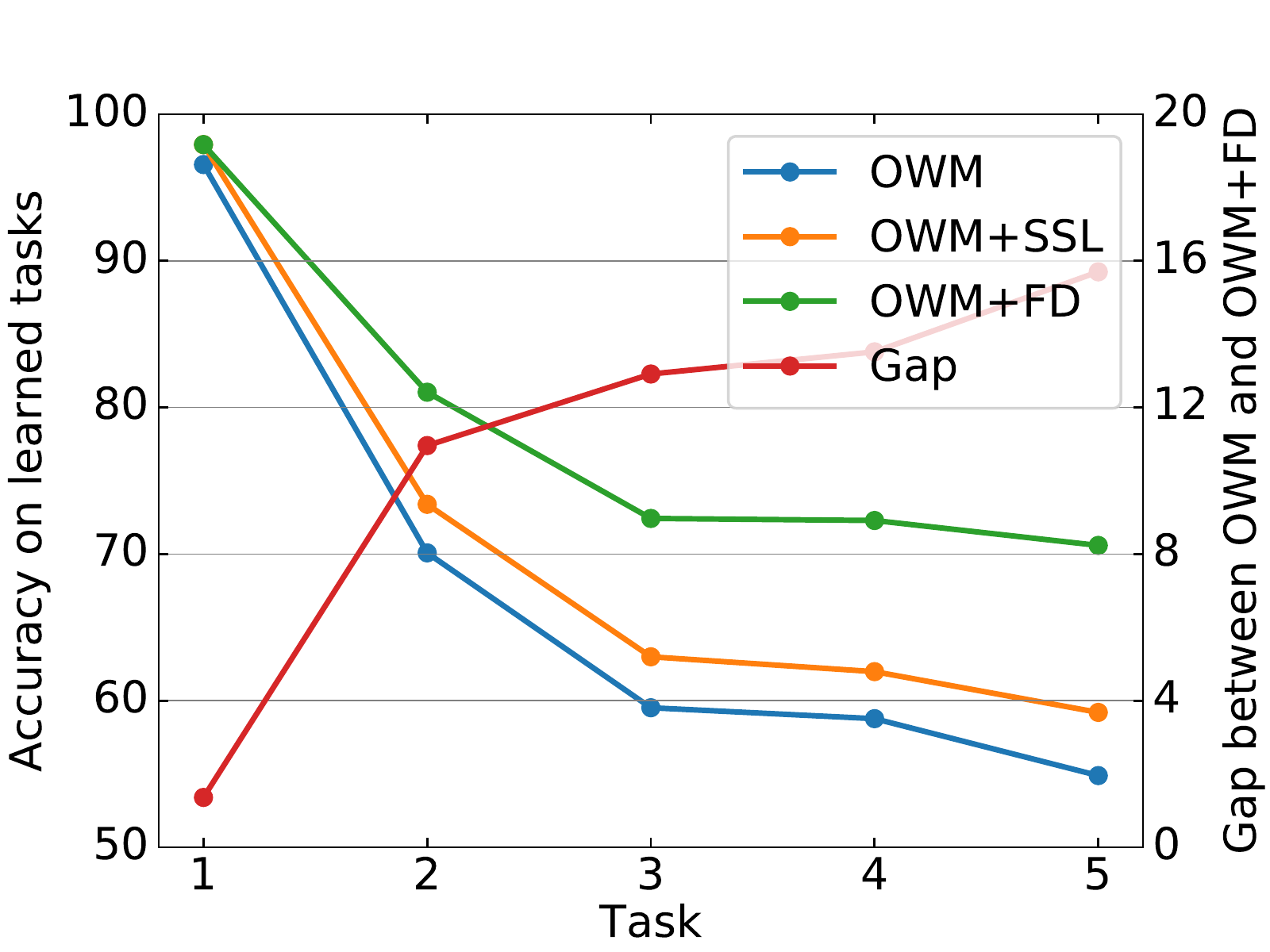}
}
\\
\quad
\subfigure[CIFAR100 (2 tasks)]{
\includegraphics[width=4cm]{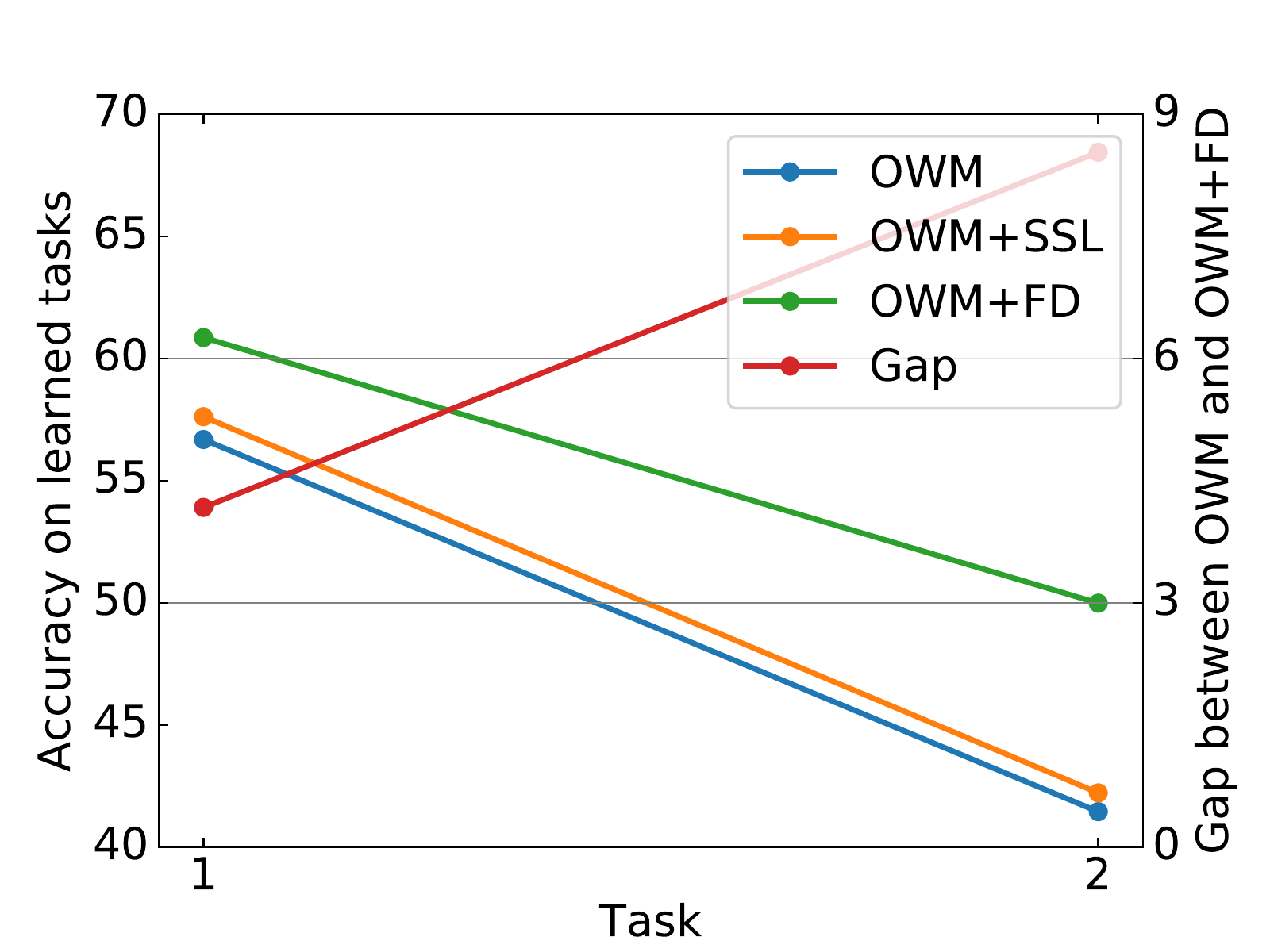}
}
\quad
\subfigure[CIFAR100 (5 tasks)]{
\includegraphics[width=4cm]{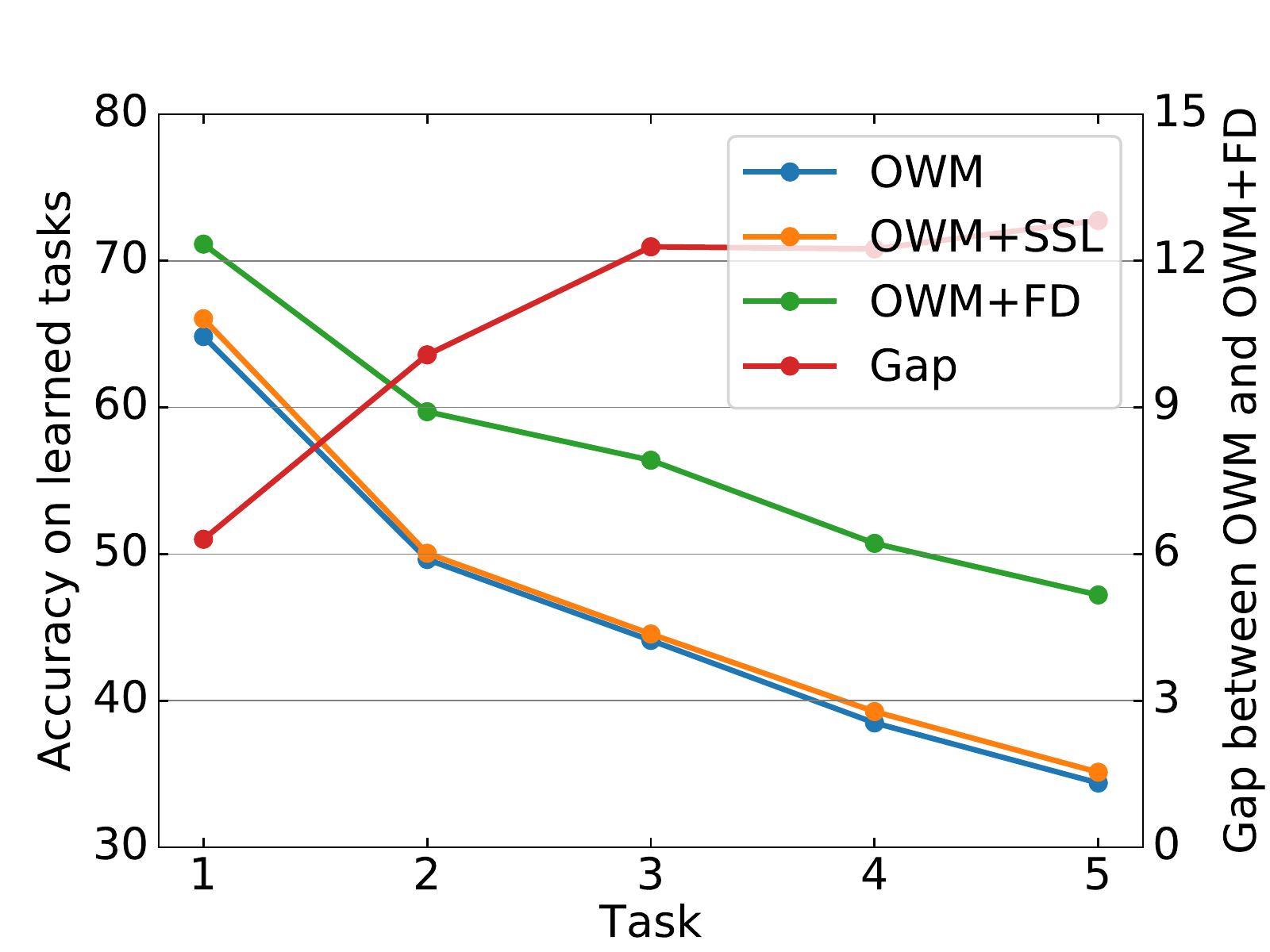}
}
\quad
\subfigure[CIFAR100 (10 tasks)]{
\includegraphics[width=4cm]{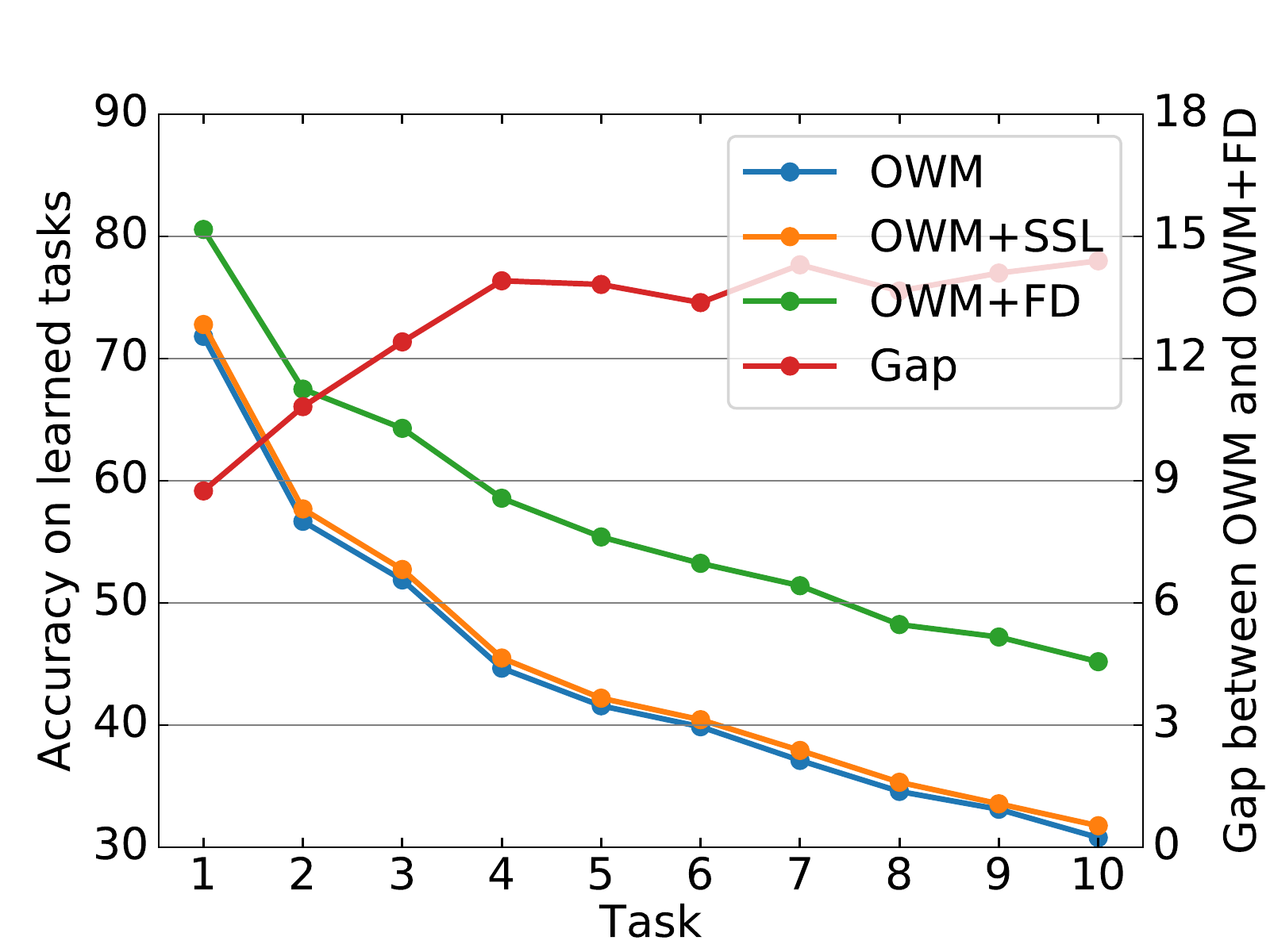}
}
\caption{Test accuracy on all learned tasks after each task.}
\label{curve}
\end{figure}

From the results on CIFAR100, it can be concluded that the disparity between OWM and OWM+FR widens as the number of tasks increases.
This is also in line with our expectations, consistent with the seriousness of PIL.
We also plot the training curve of each setting in Figure \ref{curve}, which shows test accuracy on all learned tasks after each task.
During the training procedure, the gap between OWM and OWM+FD expands gradually, as more lack of information are reflected in following tasks.
However, the rate of expanding is falling in general, as more and more features are extracted and the average amount of missing information decreases.
Above results verify our analysis of PIL in Section \ref{analysis}.

\subsection{Study of SSL}
\label{sec-e3}
In this section, we explore different ways for SSL to assist the Class-IL.
Conventionally, SSL is used for data augmentation in supervised learning.
It is not applied in our method because data augmentation usually hurts the performance on classification.
Another alternative is inspired by self-supervised data augmentation methods with aggregation (SDA+AG) \citep{Lee2019Rethinking}, which improves result of origin task.
As it may not directly apply to lifelong learning, we attempt to modify it as follow:
\begin{equation}\nonumber
L_t(x_t,y_t;\theta_t,\phi_t,\sigma_t)=\sum_{m=1}^M [L_{ce}(C(E(f_m(x_t);\theta_t);\phi_t),y_t)+\alpha_t L_{ce}(F(E(f_m(x_t);\theta_t);\sigma_t),m)]
\end{equation}
\begin{equation}\nonumber
p(y_{\leq t}|x_{\leq t},\theta_t,\phi_t)=\frac{1}{M} \sum_{m=1}^M C(E(f_m(x_{\leq t}),\theta_t),\phi_t)
\end{equation}
Besides, choice of proxy task is decisive of the improvement as discussed in Section \ref{analysis}. 
On SVHN and CIFAR10, where single proxy task can make a difference, we compare results with rotation \citep{gidaris2018unsupervised} and RGB shuffle \citep{larsson2017colorization} as transformations, combined with above OWM+SSL and OWM with modified SDA+AG, referred to as OWM+SAA, as shown in Table \ref{table-e3}.

\begin{table}[htbp]
  \caption{Results of different methods of SSL.}
  \label{table-e3}
  \small
  \centering
  \begin{tabular}{llll}
    \toprule
    \makecell[c]{Strategy} & \makecell[c]{Transformation} & \makecell[c]{SVHN} & \makecell[c]{CIFAR10} \\ 
    & & \makecell[c]{(5 tasks)} & \makecell[c]{(5 tasks)} \\
    \midrule
    \makecell[c]{OWM \citep{zeng2019continual}} & \makecell[c]{-} & \makecell[c]{$74.32 \pm 0.66$} & \makecell[c]{$54.89 \pm 0.73$} \\ 
    \makecell[c]{OWM+SSL} & \makecell[c]{Rotation} & \makecell[c]{$\mathbf{82.08 \pm 0.72}$} & \makecell[c]{$\mathbf{59.20 \pm 1.44}$} \\ 
    \makecell[c]{OWM+SSL} & \makecell[c]{RGB Shuffle} & \makecell[c]{$66.93 \pm 0.75$} & \makecell[c]{$55.36 \pm 0.42$} \\ 
    \makecell[c]{OWM+SAA} & \makecell[c]{Rotation} & \makecell[c]{$76.78 \pm 0.72$} & \makecell[c]{$56.51 \pm 0.65$} \\
    \makecell[c]{OWM+SAA} & \makecell[c]{RGB Shuffle} & \makecell[c]{$74.45 \pm 1.11$} & \makecell[c]{$55.85 \pm 1.00$} \\
    \bottomrule
  \end{tabular}
\end{table}

With each form of SSL strategies, it can be concluded that predicting rotation improves accuracy on both datasets, while predicting shuffle of RGB channels only gain benefits on CIFAR10.
This conclusion is in line with our intuition, as the former requires features of shape which overlaps with the requirement of classification, while the latter models color which has no significance for digit recognition and only increases training burden.
Under conditions that the transformation signal is useful, OWM+SSL achieves better results than base OWM and OWM+SAA, proving that it can maximize the strengths of SSL.
However, negative effect of added training burden would also be enlarged when a false signal is chosen, e.g., RGB shuffle on SVHN.
In comparison, performance of OWM+SAA is more stable.
In the improper setting, its accuracy is approximately equal with that of OWM, which means that the negative effects of SSL and data augmentation can be counteracted with the positive effect of aggregated prediction. 
While this stability also restricts the effect of SSL even though it is beneficial, compared with OWM+SSL.

\section{Conclusion, Future Work and Broader Impact}
In this paper, we propose and explore the problem of prior information loss (PIL), which has been ignored but causes significant negative effect on models for continual learning.
Our empirical results show that by mitigating this problem, the upper bound of Class-IL can be improved obviously based on current approaches.
In additional, we combine lifelong learning with self-supervised learning, and proves that it is an effective way to alleviate it.
However, there is still a huge room for improvement, as its performance relies on the selection of proxy task.
How to design or select different self-supervised learning signals remains to be a challenge.
Besides, combination of different signals is also one direction worthy exploring, which is essential for solving difficult tasks
and approximating the upper bound.
Finally, this work does not present any foreseeable societal consequence.

\end{document}